%% file: sample-sigconf.tex
\documentclass[sigconf]{acmart}

\usepackage{booktabs} 
\usepackage{graphicx}
\usepackage{subfig}
\usepackage{blindtext}
\usepackage{dblfloatfix} 
\usepackage{amsmath}

\copyrightyear{2018}
\acmYear{2018}
\setcopyright{acmcopyright}
\acmConference[CF '18]{CF '18: Computing Frontiers Conference}{May 8--10,2018}{Ischia, Italy}
\acmBooktitle{CF '18: CF '18: Computing Frontiers Conference, May 8--10, 2018, Ischia, Italy}
\acmPrice{15.00}
\acmDOI{10.1145/3203217.3203282}
\acmISBN{978-1-4503-5761-6/18/05}

\acmArticle{4}

\begin{document}
\title{QUENN: QUantization Engine for low-power Neural Networks}

\author{Miguel de Prado}
\affiliation{Integrated Systems Laboratory, ETH Zurich}
\affiliation{He-Arc Ingenierie, HES-SO}
\email{miguel.deprado@he-arc.ch}
\author{Maurizio Denna}
\affiliation{Nviso}
\email{maurizio.denna@nviso.ch}
\author{Luca Benini}
\affiliation{Integrated Systems Laboratory, ETH Zurich}
\email{lbenini@iis.ee.ethz.ch}
\author{Nuria Pazos}
\affiliation{He-Arc Ingenierie, HES-SO}
\email{nuria.pazos@he-arc.ch}

\renewcommand{\shortauthors}{M. de Prado et al.}

\begin{abstract}
Deep Learning is moving to edge devices, ushering in a new age of distributed Artificial Intelligence (AI). The high demand of computational resources required by deep neural networks may be alleviated by approximate computing techniques, and most notably reduced-precision arithmetic with coarsely quantized numerical representations. In this context, Bonseyes comes in as an initiative to enable stakeholders to bring AI to low-power and autonomous environments such as: Automotive, Medical Healthcare and Consumer Electronics. To achieve this, we introduce LPDNN, a framework for optimized deployment of Deep Neural Networks on heterogeneous embedded devices. In this work, we detail the quantization engine that is integrated in LPDNN. The engine depends on a fine-grained workflow which enables a Neural Network Design Exploration and a sensitivity analysis of each layer for quantization. We demonstrate the engine with a case study on Alexnet and VGG16 for three different techniques for direct quantization: standard fixed-point, dynamic fixed-point and k-means clustering, and demonstrate the potential of the latter. We argue that using a Gaussian quantizer with k-means clustering can achieve better performance than linear quantizers. Without retraining, we achieve over 55.64\% saving for weights' storage and 69.17\% for run-time memory accesses with less than 1\% drop in top5 accuracy in Imagenet.
\end{abstract}

%
%



\maketitle

\input{samplebody-conf}

\bibliographystyle{ACM-Reference-Format}
\bibliography{sample-bibliography}

\end{document}

%% file: samplebody-conf.tex
\section{Introduction}
Deep Learning (DL) has evolved rapidly in the last few years thanks to the boost of hardware accelerators such as GPUs or clusters of CPUs \cite{25}. The increase of computational power has allowed networks to grow deeper and wider \cite{26, 27} to greatly overpass human accuracy level \cite{28}. Convolutional Neural Network (CNN) architectures are a class of DL which have spread across a large variety of tasks and applications in the domain of computer vision and speech recognition \cite{29, 30}.
\par
Remarkable accuracy comes with the cost of power consumption and large memory footprint. While CNN architectures have dramatically conquered cloud services, such as Google \cite{31}, Amazon \cite{32}, they have not yet completely flourished for embedded or edge devices due to the vast computing resources they need \cite{49}. There exist several techniques to reduce computational requirements and memory consumption to make CNN suitable for embedded devices, i.e. pruning \cite{19,21}, quantization and compression \cite{16, 17}. The application of these techniques degrades the accuracy of CNN implementation which can be partially solved by retraining or fine-tuning the network. However, setting up a training environment is not trivial due to the number of algorithms and parameters that need to be tweaked. \par
Techniques that aim to reduce numerical precision, e.g. quantization, are bound to hardware processor architectures which only support certain data types of underlying computing units, e.g. 32-bit floating point or 16-bit fixed point. This fact prevents small stakeholders from using smaller or rare data types with 10, 8 or 6 bits, which can achieve the similar performance as floating point, in specific cases. Only large companies, such as Nvidia, can overcome this situation and lead the DL market thanks to their expertise in building end-to-end systems and hardware processors or accelerators to make neural networks more efficient. However, acquiring specific hardware for specific use cases usually involves using their own proprietary libraries, which constrains stakeholders to use them openly. \par
In contrast to monolithic and proprietary sources systems, we form part of Bonseyes \cite{33, 41}, an European collaboration to bring Deep Learning to any stakeholder. The initiative makes available an AI Marketplace where AI services, tools, data and knowledge can be found and exchange. Deep Neural Network (DNN) tools focus especially on reducing development time and optimizing deployment on embedded, constrained and distributed systems. One of the main goals of Bonseyes is to generate efficient and tunable code for a wide range of heterogeneous embedded and IoT systems, maximizing the portability of CNNs through a comprehensive neural network framework called LPDNN. LPDNN, originally based on CaffePresso \cite{40}, also puts into practice efficient approximation computing techniques such as quantization, pruning \cite{41}, to reduce the memory footprint and improve data traffic. \par
Traditionally, approximation has been applied generally over CNN architectures without discerning between the various kinds of layers \cite{10} and only takes into consideration the final accuracy of the network under well-known datasets like Imagenet \cite{34}. There is a lack of analysis in the internal behavior of networks when, for instance, quantization is applied to a single layer or group of layers and their position within the network.  \par
The objectives of this work are twofold: \textit{i)} to shed light on the internal behavior of a network when different techniques of quantization are applied; and \textit{ii)} to provide an introspective analysis of the main layers of a network: Convolution, Pooling, Relu and Fully Connected in comparison to a baseline 32-bit floating-point implementation. Both objectives aim to curtail computational requirements and memory storage for deployment of CNN on embedded devices. \par
To fulfil these objectives, we present LPDNN and more specifically, an engine which integrates a modular quantization pipeline. The quantization pipeline analyzes and selects an optimum implementation with minimum memory footprint while keeping high accuracy. To validate the engine, a fine-grained analysis of direct quantization and its influence on the layer and its relative position is demonstrated for Alexnet \cite{26} and VGG16\cite{48}. Our approach without retraining is complementary to those implementing retraining and does not need massive computational resources and expertise that retraining processes need, turning them impractical for many users. \par
The paper is organized as follows: In Section 2, the State-of-the-Art is presented.  Section 3 describes LPDNN, its architecture and the quantization workflow. In section 4, various kinds of quantization and the methodology pursued are detailed. Section 5 shows the results and analysis that have been obtained. Finally, in section 6, the concussion and future work is expressed.
\section{Related work}
In the literature, a large variety of works aim to reduce computational effort and memory footprint: \textit{i)} Pruning is a method whereby reducing the number of connections between neurons, sparse matrixes are obtained which reduce matrix multiplication effort \cite{18, 20}. \textit{ii)} Compression of a network can also be achieved by applying low-rank tensor decomposition and fine-tuning to the convolution kernels at the minimum loss \cite{15, 16}. \textit{iii)} Quantization, a method to reduce the numerical precision of variables, has been studied extensively for CNNs, especially coupled with retraining of weights and activations (outputs of the layer) with low precision \cite{3, 5, 6, 14} to the very extreme of reducing numerical precision down to binary \cite{4, 7}. There has been a wide range of works related to optimization techniques to minimize the quantization error: L2 \cite{1, 8} and SQNR \cite{2}. In \cite{1}, Anwar et al. also studied the effect of quantization to improve sparsity in a network.
\par
Fixed-point quantization, in particular dynamic fixed-point quantization, has been proposed in some works such as \cite{10}, and specifically for CNNs in \cite{2, 11} to improve accuracy by adapting the fractional length of the weights and activations independently from the distribution range of the layers. In \cite{13}, Qiu et al. implemented a greedy algorithm to find the optimal fractional length per layer while in \cite{11}, the integer length is first fixed to cover the maximum weight's or activation's value of the layer. The fractional length is then selected as a subtraction of the integer length from the bit width of the variable. Shafique et al. \cite{47} provide a deep analysis of hardware-oriented quantization aimed to improve accelerators energy efficiency. To do so, they show an Evolutionary Circuit Approximation to find a suitable solution to the quantization problem. They demonstrate that different input data does not change the data distribution of the feature maps, leaving the quantization resilience on the network itself. Besides, the authors prove that data path's distribution has a major impact on the network as certain feature maps are more sensitive to quantization than others in the same layer. However, none of these methods studies the effect of k-means clustering \cite{22} for optimum quantization. \par
K-means clustering involves forcing variables belonging to a certain interval to be identical and take just one common value. In \cite{21}, Han et al. proposed to add this technique on top of pruning before a retraining phase achieving 35x reduction of weights and no loss of accuracy. However, the technique is only applied to weights and not to activations, leaving them at 32 bits. Further, there is no analysis of applying the method to only a subset of layers nor the influence that such layers may have within the network's data path. In contrast, in this work, we propose k-means clustering for weights and activations independently as well as an analysis for each layer and the influence of their data distribution on the network's performance. \par
In most of previous works, for instance, in \cite{1, 11}, the effect of dynamic fixed-point is only studied for Convolutional (Conv) and Fully Connected (FC) layers without any analysis of their position within the network. Layer's tolerance to quantization is partially analyzed in \cite{1}, where the authors show that FC layers, especially the last one, are more sensitive than Conv layers to quantization. Furthermore, analysis is only performed on small networks for the Cifar-10 \cite{35} and MNIST \cite{36} datasets. Judd et al. apply in \cite{43} quantization to only a single layer at a time to demonstrate how large networks' internal behavior and accuracy vary when different layers of the same network are quantized. They propose an algorithm to find the suitable bit width, fractional and integer length for all layers. \par
In \cite{44}, Lai et al. show that the range of weights is the main factor that affects the accuracy of a network. They propose to represent convolutional and fully connected weights in floating point format while keeping activation in fixed point and demonstrate an efficient multiplier for a mixed fixed-point floating-point multiplication. We demonstrate in this work, that the range can be fully represented with fixed point format by using k-means clustering and that Gaussian distribution quantizers can reduce the influence of large-range distributions.  \par
Nvidia has recently released a new version of TensorRT \cite{45}, a framework for quantization of neural networks where 32-bit implementations are converted into 8 bits. To achieve a wider range than the one given by 8 bits [127, -128], they multiply the 8-bit range by a 32-bit floating-point scale factor which can be adapted per layer. They show that saturation of activations' values does not harm the performance by finding an optimum saturation threshold through minimization of the KL-divergence. Nvidia's work is implemented, as proprietary libraries for GPUs which perform transparently to the user without the possibility to fine-tune it for a specific application or hardware architecture. There is no analysis for quantization to lower number of bits, e.g. 4 or 6, which could be implemented in FPGA and the knowledge obtained from quantization is not published. In our work, we show that linear k-means clustering using fixed-point format achieves a suitable range without the need of using floating-point multipliers and that the use of a Gaussian quantizer, relates to saturating the activations as proposed by Nvidia.

\section{LPDNN}
To better understand the proposed quantization workflow and its tight integration within the inference engine, a detail description of LPDNN will be presented. LPDNN (Low Power Deep Neural Network) is an inference engine developed within the Bonseyes project for Deep Learning. LPDNN is an enabling framework which provides the capabilities and tools to produce portable implementations of neural networks efficiently for constrained and autonomous AI applications such as: Automotive Safety and Cognitive Computing, Consumer Emotional Virtual Assistant and Healthcare Patient Monitoring.
\subsection{Architecture}
\begin{figure}
\centering
    \includegraphics[width=0.45\textwidth, height=2.12in]{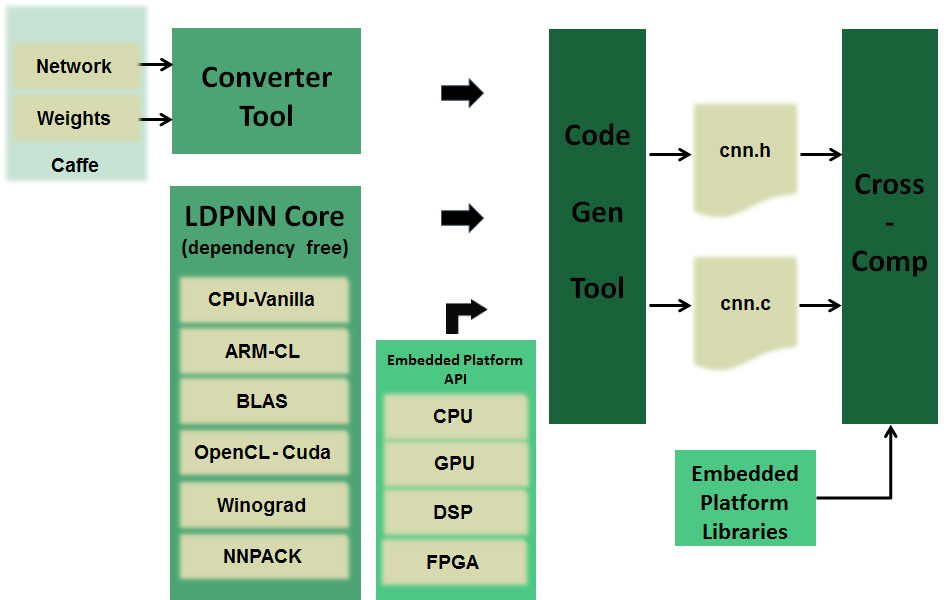}
	\caption{LPDNN Architecture. On the left, Converter Tool from Caffe, LPDNN Core and the computing libraries that may be plugged in. In the middle, available platforms APIs and Code Generator Tool. On the right, cross-compilation and linking to specific embedded platform.}
    \label{fig:arch}
\end{figure}
LPDNN presents a modular architecture where any given Caffe \cite{37} network description and pre-trained model can be parsed automatically by a conversion tool to generate an executable representation. As it can be seen in Fig. \ref{fig:arch}, LPDNN contains a Neural Network Core which embraces a set of dependency-free functions implemented in ANSI C that supports inference of neural networks on any embedded device. The core module is complemented by a set of plugins that can be built together with the core to produce optimized code for a specific computing system. Each plugin can make use of other available computing libraries such as OpenBlas, MKL, NNPACK, cuDNN. 
The plugin-based architecture allows the main core to remain small and dependency-free while additional libraries are only included when needed and for specific platforms, notably increasing the portability across systems. Besides, cross-compilation and specific tools are added to support a wide range of heterogeneous computing platforms such as CPUs, GPUs, DSPs, VPUs. The core is able to support multiple plugins at the same time; this flexibility makes it easy to write and experiment with optimized algorithms for some specific layer types and to dispatch the execution of each layer to the most suitable implementation according to the network architecture, target platform and desired accuracy and performance specification. \par
The previous goals can be achieved by providing input directives to the Code Generation Tool, to specify the plugin, computing library and the approximate computing strategy to be applied for each layer. For example, convolutions may run on the GPU by cuDNN while the other layers remain in the CPU. By these means, efficient C code is generated which is then cross-compiled and linked to the specific embedded platform libraries.

\subsection{Quantization workflow}
\begin{figure}
\centering
    \includegraphics[width=0.45\textwidth, height=1.8in]{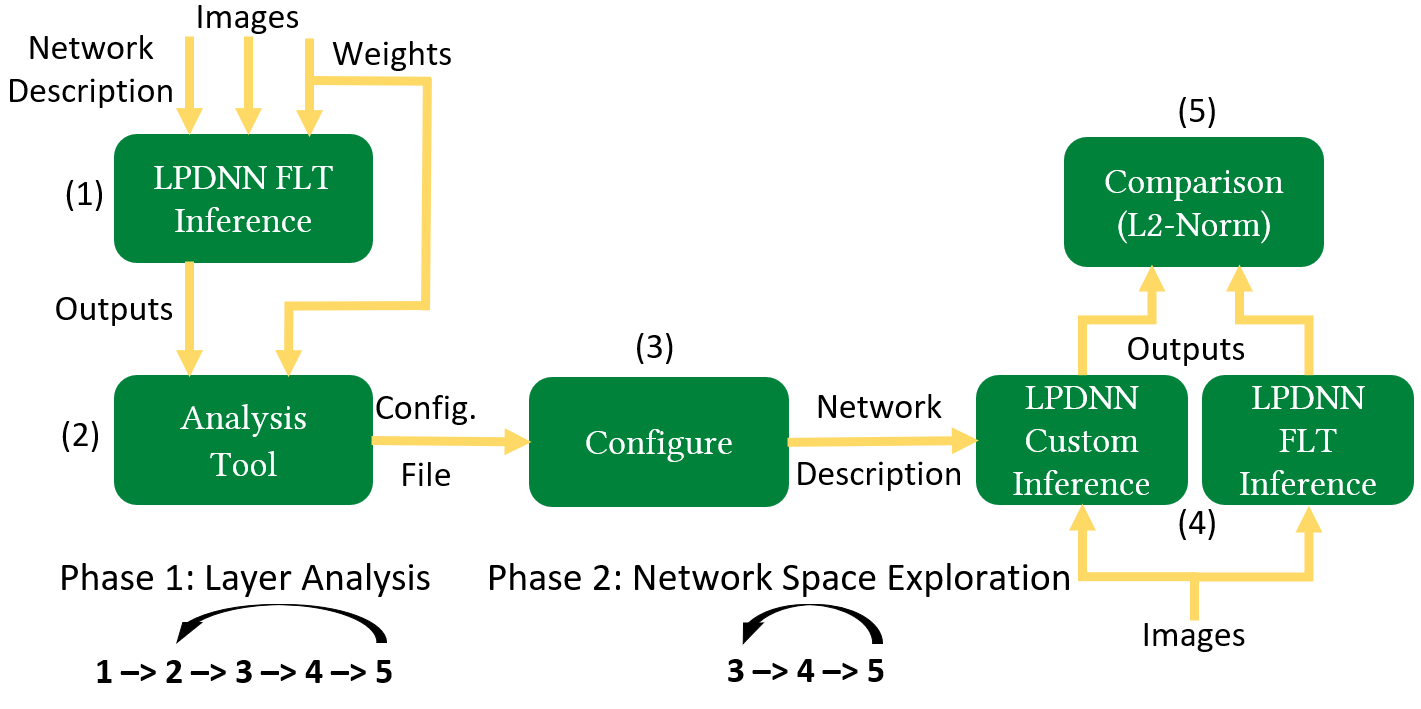}	
	\caption{Quantization pipeline in LPDNN is made up of two phases. First phase corresponds to points Layer Analysis and loops over steps 2-5. Second phase corresponds to Network Space Exploration and loops over steps 3-5.}
    \label{fig:work}
\end{figure}

Quantization can be fine-grained implemented and customized in LPDNN through an iterative pipeline. A single layer and variable type, i.e. weights or activations, can be quantized while keeping all other layers in single-precision (32-bit) floating-point format (FLT) to analyze the influence of quantization techniques and the sensitivity of a layer or group of layers within a network. The pipeline process allows the selection of the number of iterations for each parameter search to trade-off between the analysis time and performance. As it can be seen in Fig. \ref{fig:work}, the pipeline contains two main phases: 1) Layer Analysis and 2) Network Space Exploration.

\begin{figure*}[ht]
\centering
    \includegraphics[width=0.8\linewidth, height=3.5cm]{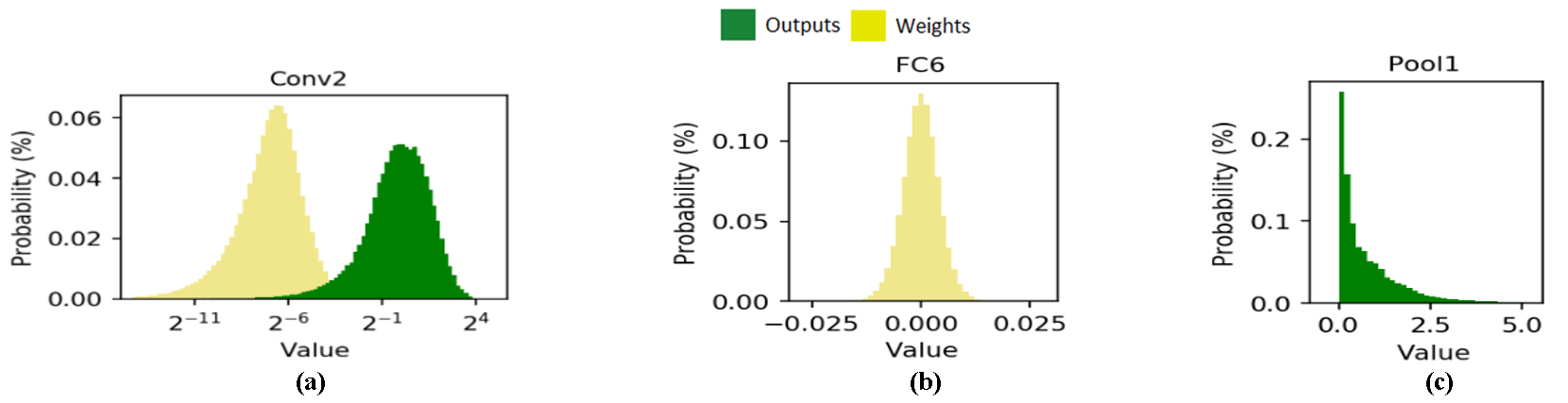}	
	\caption{Alexnet Analysis: a) Log histogram of weights and activations for automatic calculation of Fractional and Integer Length from the Analysis Tool. b) Gaussian distribution of weights. c) Distribution suited for a linear quantizer.}
    \label{fig:distri}
\end{figure*}

\begin{figure*}[ht]
  \centering
  \subfloat[]{\includegraphics[width=0.4\linewidth, height=4cm]{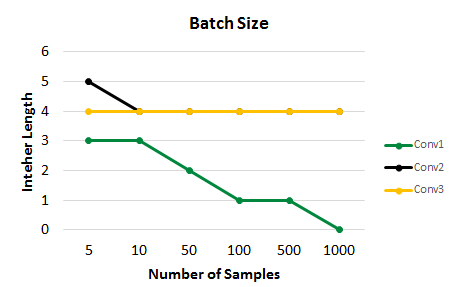}\label{fig:ff1}}
  \hspace{2cm}
  \subfloat[]{\includegraphics[width=0.4\linewidth, height=4cm]{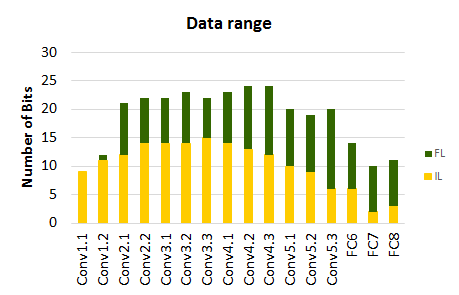}\label{fig:ff2}}
  \caption{: a) Batch size impact in first layers of Alexnet. b) Necessary bit width [IL, FL] to cover all data range of VGG16.}
  \label{fig:data}
\end{figure*}

In the Layer Analysis phase, the quantization technique is selected e.g. dynamic or k-means. The objective of this phase is to search for the distribution details and optimum fixed-point parameters, e.g. Integer and Fractional Lengths [IL, FL], for several bit widths, see Fig. \ref{fig:distri}-a. In order to minimize the quantization error, the search is performed automatically for activations and weights independently in each layer. First, the Analysis Tool collects the LPDNN network description, the pre-trained model's weights and calculates the distribution details of the latter (min, max, mean, STD, FL and IL), see Fig. \ref{fig:distri}-a. The distribution details of the network's activations should be calculated by inferring a batch from a subset of the validation set that is able to represent the dataset distribution. Large batches are very costly and time consuming and mainly affect just the first layer of the network, see Fig.  \ref{fig:data}-a (in this work,we have selected a batch of 50 images).
\par
All the quantization parameters are written onto the configuration file which is used to infer and compare the Custom and FLT implementation. Although, a first IL and FL have been selected from the distribution in the Analysis Tool, it does not mean that they are the optimum values for a CNN implementation, especially for k-means technique (see Section 4 for more details on the quantization process). To achieve the values that minimize the quantization error, an iterative process is applied over steps 2-3-4-5 for a fixed range of FLs [8-20], which has been experimentally proven to perform better. Finally, the best configuration is selected, see Fig.\ref{fig:work}. \par
In the second phase, the Network Space Exploration is carried out taking as input the configuration file with the optimum parameters per layer from the Layer Analysis phase. The objective of the Network Space Exploration is to determine the sensitivity of each layer to bit-width scaling and the set of possible configurations when a group or all layers of the network are quantized. This is achieved by iterating over the following steps 3-4-5: Configure, Inference \& Comparison, see Fig.\ref{fig:work}. \par
The configuration step takes in the configuration file and selects the layer/s and the variable type to quantize, e.g. weights and/or activations of Conv1, and creates a network description. Next, in step 4, the custom and the standard FLT networks are inferred independently over the batch of images. In step 5, the activations of the FLT and custom inferences are compared against each other and the Frobenius Norm distance (here stated as L2-Norm for matrices) between the FLT and custom implementation is calculated for every layer and image following Eq. 1, where \textit{x} is the pixel in row \textit{i}, column \textit{j} and channel \textit{k}. All L2-Norm distances corresponding to each image of the batch are averaged to give a unique result. At last, the configuration which minimizes the quantization error is selected.
\vspace{-0.2cm}
\begin{equation}
  L2Norm = \bigg[ \sum_{k,i,j}^{c,m,n}\parallel x_{i,j,k} - x'_{i,j,k} \parallel^2\bigg]^{1/2}
\end{equation}
\vspace{-0.5cm}
\section{QUANTIZATION METHODOLOGY}
Quantization aims to reduce the memory storage and traffic of an implementation by decreasing the number of bits of a variable and therefore its numerical precision. Standard single-precision variables (32-bits) can be substituted by variables with fewer bits which degrades the precision or range that such variable can represent, e.g. IEEE 754 32-bit floating-point \citep{38} max. and min. normalized representable values are 
\begin{math} 
	\approx 2^{127}
\end{math} 
and
\begin{math} 
	2^{-126}
\end{math}  
respectively. Having the same format with 16-bit variables, max and min are 
\begin{math} 
	\approx 2^{15}
\end{math}  
and
\begin{math} 
	2^{-14}
\end{math} 
while taking up half of the memory. Therefore, a trade-off between range, accuracy and memory usage must be made depending on the constraints of the system and application.  \par
\begin{figure}[hb]
\centering
    \includegraphics[width=0.3\textwidth, height=0.5in]{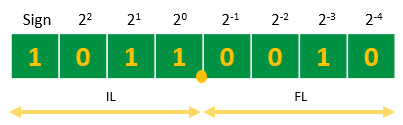}	
	\caption{Fixed-point representation.}
    \label{fig:fix}
\end{figure}

While floating-point format counts on a dynamic range where the fractional and integer part may vary, fixed-point format fixes the Integer and Fractional Length [IL, FL] \citet{11}. Fixed-point format is signed and can be represented as shown in Fig.\ref{fig:fix}. Although fixed-point format's precision decrease notably compared to floating point format, fixed-point multiplications are faster and less hardware demanding in terms of power and area than floating-point multiplications \citep{39}. Therefore, forcing a fixed-point format is a common technique in embedded systems where floating-point multipliers are considered to be expensive. \par
In this work, dynamic fixed-point quantization has been implemented in LPDNN. In addition, k-means clustering \cite{42} has also been integrated and both will be compared.

\subsection{Dynamic fixed point and k-means clustering}
Standard fixed point involves using the same integer and fractional length for all variables which reduces considerably the efficiency of the implementation of neural networks since each layer and variable, weights or activations, contains different distribution range. This work has been inspired by \cite{11} where an implementation for dynamic fixed-point is shown. In our work, all layers have independent quantization parameters among them, e.g. IL, FL, etc., as well as within the same layer, forming weights and activations two independent groups. The variables within these two groups do share the same quantization parameters. \par
The proposed approach to select the dynamic fixed-point quantization parameters is as follows: the IL is fixed in first place to cover the integer range of the data, see Eq. 2, and the FL in second place depending on the IL and the variable's bit width (BW), see Eq. 3. For high bit widths, there might be an over saturation due to a high FL value and therefore an automatic iterative approach has been followed to detect Y, the optimum FL per layer. The iterative approach runs for a fixed-range of possibilities which have been experimentally determined. In Eq. 3 it can be seen that Y depends on the variable V, either weights or activations of a layer to be quantized.
\begin{equation}
\begin{aligned}
 \hspace{0.9cm} IL = \lceil \log{(\max{(V)})}\rceil
 \end{aligned}
\end{equation}
\vspace{-0.4cm}
\begin{equation}
\begin{aligned}
  FL = BW - IL \\
  If FL > Y(V):\\
  \ FL = Y(V)
\end{aligned}
\end{equation}
The k-means clustering approach, implemented in the current work, has been inspired in \cite{21}, where Han et al. implemented Shared Weights. Han et al, cluster weights of CNNs into Kintervals in order to minimize the clustering error per layer:
\begin{equation}
  \arg_{s} \min \sum_{i=1}^{k}\sum_{x\in S_{i}}\parallel x - \mu_{i} \parallel^2
\end{equation}
where \begin{math}\mu_{i}\end{math} is the shared weight within the interval and x the real value. By these means, weights do not longer contain a particular value but an index of a table in which all known intervals' value are stored. Therefore, the memory saving is significant since each weight gets reduced as in Eq. 5:
\begin{equation}
  \log_{2}(Kintervals)/ \textrm{Standard bit width}
\end{equation}
to which we only need to add up the table's size: \textit{Kintervals * bit width} (of the shared variable). In this work, we also apply the technique to activations and we use fixed-point representation with suitable FL, instead of floating-point data type. Besides, the number of intervals or clusters is defined as a power of two of the bit width, which simplifies the search for stakeholder that do not possess the tools and computational power to undertake such operations and we will show that this simplification does not harm the final accuracy.
\vspace{-0.2cm}
\vspace{-0.2cm}
\subsection{Layer quantization}
Quantization of a layer can be carried out independently of other layers, i.e. one layer is quantized to 4-bit, a second to 8-bit while the others remain with 32 bits. For each layer, activations and weights (if applicable) can be quantized to any bit width\footnote{The value can represent any number of bits although the data type remains in 16 bits, e.g. max value that a 5-bit variable may take would be 00000000 00011111} either in dynamic fixed-point or k-means clustering format. Both variable types must have the same format. \par

Fixed-point quantization is carried out by multiplying the FL, obtained as in Eq. 3, to the floating-point variable in order to include FL number of fractional bits into an integer number. The FLs are all power of two and therefore, this multiplication is implemented by a shift. Finally, a deterministic rounding (round-to-nearest) \cite{6} is applied based on the decimals that are going to be left over and not taken by the fixed-point implementation. The conversion from fixed point to floating point is performed in the same manner but dividing by FL instead of multiplying.\par
The implementation of k-means clustering in LPDNN is carried out by an optimal quantizer \cite{23, 24}. In this work, a uniform quantizer for linear (L) and Gaussian (G) distributions has been implemented, that is, a quantizer whose step size or intervals are all equal over the range provided by the layer distribution type. To quantize a layer, the following steps must be followed: 
\begin{enumerate}
\item Set the data range limits by collecting the min/max (L) or mean/variance (G) of the layer from the configuration file.
\item Normalize the data of the group to be quantized to [0,1].
\item Multiply unit-range data by the given number of cluster or intervals so that all data falls into a certain interval. All values within an interval will no longer possess their original value but the index of the interval.
\item Create an optimum representative table of values, based on the distribution (linear or Gaussian), range and bit width, where all mid center values of each cluster are stored.
\item Map the data indexes onto the representation table so that the shared value of the interval is obtained.
\end{enumerate}

\begin{figure*}[ht]
\centering
    \includegraphics[width=\linewidth, height=1.3in]{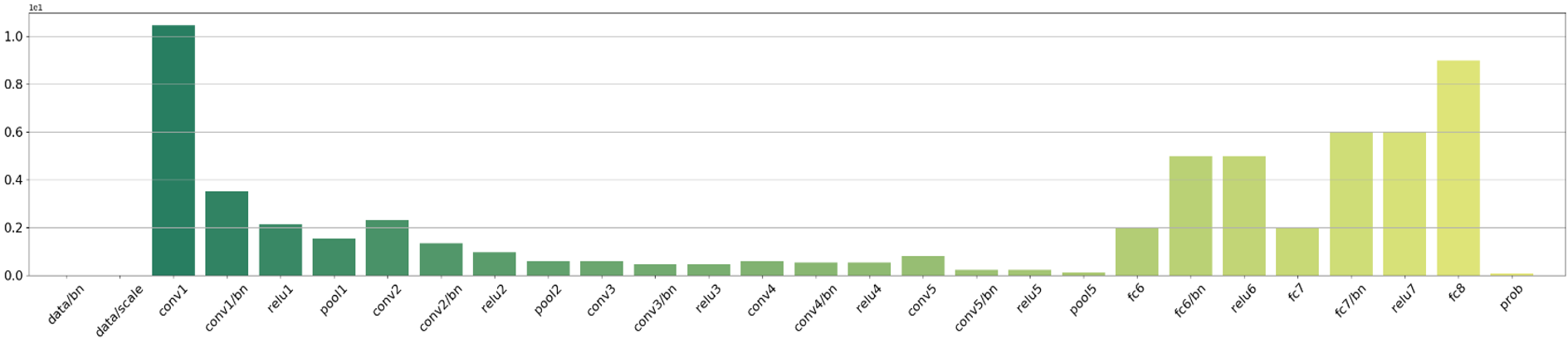}	  \caption{Example of single-layer quantization for Alexnet. Conv1's activations, which are optimally computed together with Relu1 and BN1, have been quantized to 6 bits while keeping all other layers at 32-bit floating point.}
    \label{fig:alexnet}
\end{figure*}
\begin{figure*}[ht]
  \centering
  \subfloat[]{\includegraphics[width=0.33\linewidth, height=4cm]{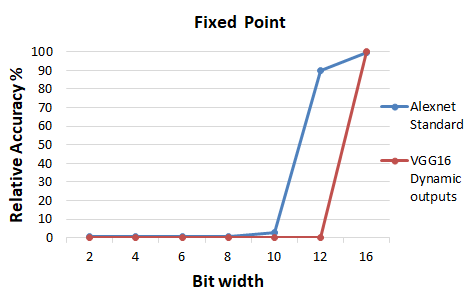}\label{fig:f1}}
  \subfloat[]{\includegraphics[width=0.33\linewidth, height=4cm]{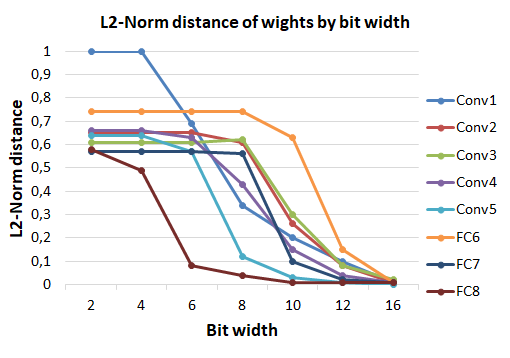}\label{fig:f2}}
  \subfloat[]{\includegraphics[width=0.33\linewidth, height=4cm]{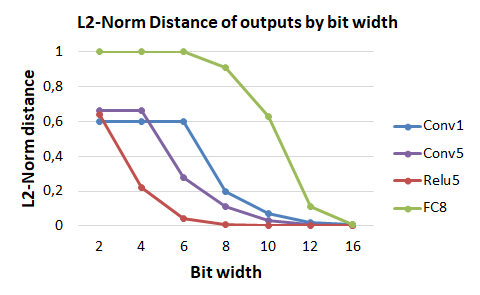}\label{fig:f3}}
  \caption{: a) All figures Y-axis contain relative values compared to the FLT baseline work. a) Y-axis represents the fixed-point accuracy when changing the bit width homogeneously. b) Y-axis represents the final L2-Norm distance for Alexnet when varying the weights' bit width, the lower the better.  c) Same Y-axis as b when varying the activations' bit width.}
  \label{fig:layer}
\end{figure*}

\section{RESULTS AND ANALYSIS}
LPDNN offers custom quantization where the user can analyze and optimally select the layers that less sensitivity or loss of accuracy exhibit to quantization while saving the most possible memory space and computation time. As an example, we present an analysis for the Convolution, ReLu, Pooling and Fully Connected layers for AlexNet and VGG in one of the largest and most challenging datasets, Imagenet. First, we carry out a Layer Analysis to obtain the optimum per layer parameters. Next, we perform a Network Space Exploration by quantizing one layer at a time for either activations or weights and then both together while changing the bit width. Afterwards, we quantize a group of layers and compare the different techniques. Finally, we choose a custom and optimum implementation for the whole network given by the quantization workflow. We have used Alexnet with Batch Norm \cite{46} and VGG16 \cite{48}, whose top1 and top5 accuracy are: [60.1\%; 81.9\%] and [73.0\%; 91,2\%] respectively in the validation set. We also consider each convolution, pooling and Relu layer independently, e.g. the implementation of Alexnet contains 28 layers as in Fig. \ref{fig:alexnet}. The results of the analysis and the comparisons are all relative to the 32-bit floating-point baseline implementation and the memory saving for both weights' storage and activations' memory traffic, i.e. activation tensor size that is passed through the network layer by layer, are theoretically calculated since no ASIC implementation has been developed.
\subsection{Single-layer quantization}
In this section, we show an analysis of the internal behavior of Alexnet by quantizing specific layers and variables. Quantization of a single layer in the whole network may seem trivial but a quick analysis can reveal valuable information. In table 1 and 2, we can see the size of weights and activation per layer. Clearly, FC6 accounts for most of the weights (61.92\%). By quantizing the weights of FC6 to 6 bits using k-means clustering, we can achieve a reduction of 50\% in memory size for weights' storage with no loss in top5 and only a drop of 2\% in top1 accuracy. Likewise, Conv1 makes up for about 39\% of the activations' memory traffic. We accomplish a 32\% overall reduction in memory traffic when quantizing the activations of Conv1 to 8 bits with just 1\% drop in top1 and top5 accuracy, see Fig. \ref{fig:alexnet}. 
The sensitivity of the layers to quantization varies across a network mainly due to three factors: \textit{i)} the range of the data, \textit{ii)} the range and precision of the variables that hold the data and, \textit{iii)} the position of the layer. The first two factor are highly correlated as we can see in Fig. \ref{fig:layer}-b. The weights of FC6 quickly degrades compared to the floating-point baseline as the bit width is reduced since FC6 needs 14 FL bits to represent all the small weight values. Likewise, quantizing Conv1 brings a higher penalty than any other convolution layer as the data range of a real image in the input is higher than the abstract features deeper in the network, see Fig. \ref{fig:layer}-c. Although, range plays a key role, the precision and the position of the layer become a crucial factor in FC8 since it is the last computing layer. Although in average, the range of FC8's activations could be represented with 6 bits, the penalty in missing small and precise values is huge as a totally different class could be selected in the high-top classification ranks.
\begin{figure}
\centering
    \includegraphics[width=0.47\textwidth, height=0.5in]{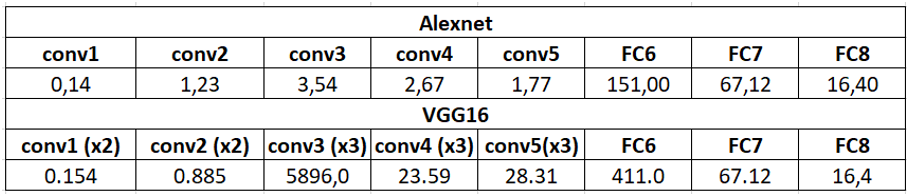}
    \captionof{table}[foo]{Weights of baseline Alexnet and VGG16 in Mbytes. BN and scale weights of Alexnet are negligible.}
    \label{fig:w_size}
\end{figure}
\begin{figure}
\centering
    \includegraphics[width=0.47\textwidth, height=0.5in]{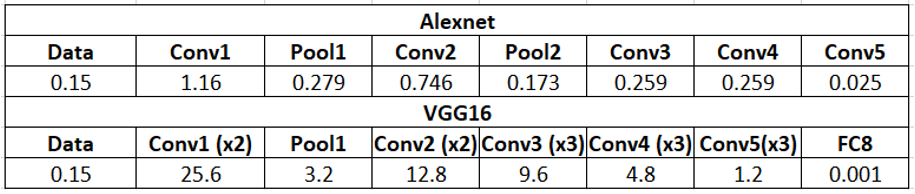}
	\captionof{table}[foo]{Largest baseline activations from Alexnet and VGG16 in MB. Relu and BN are optimized and combined with Conv layers to compute the activations only once.}
    \label{fig:o_size}
    \vspace{-0.7cm}
\end{figure}
As shown in Fig. \ref{fig:alexnet}, the L2-Norm distance of the network varies internally. The height represents the distance to the baseline implementation which we take as ground truth. The last layer gives out the distance between [0, 1], being 0 identical and 1 completely wrong. Overall, Relu and pooling layers are the most quantization-friendly layers as they partially recover the quantization error since the former only get the positive range of the variables and the latter minimize the range by selecting the max. over a window. In general, weights need more bits than activations to achieve the same accuracy. Weights of FC layers are less sensitive than those of Conv layers while the activations of both types of layers depend more on the position of the layer.
\subsection{Quantization techniques}
In this section, we show the analysis and point out the differences between standard fixed-point, dynamic fixed-point and k-means clustering quantization for both Alexnet and VGG16.
\subsubsection{\textbf{Standard fixed point}}
This technique involves using a uniform configuration where all layers have the same numerical representation (FL, IL, BW). In Fig. \ref{fig:layer}-a, we show how Alexnet's accuracy decreases when the BW is lowered. As FC6 weights and FC8 activations need a higher bit width to represent the range, they become a bottleneck for the network and the accuracy quickly drops after 12 bits. This finding is consistent with the work in \cite{43} which also shows that 16-bit representation have minor impact in the accuracy. VGG represents a higher challenge than Alexnet for quantization as it contains more number of layers and their feature maps' depth and size increases, see table 2. Besides, it does not use Batch Norm layers which implies that the distribution data of the layers' activations is spread over a broader range and it easily overflows small data types, see Fig. \ref{fig:data}-b. This fact makes difficult to quantize both weights and activations as there is a need of a higher number of bits to represent the data and dynamic FL. 
\subsubsection{\textbf{Dynamic fixed point}}
In this section, we prove that dynamic fixed point achieves better performance than standard fixed-point due to the adaptability of the quantization parameters. Since Alexnet's FC8 activations are the most sensitive to quantization, we leave it 16 bits while we shrink Conv5, FC6 and FC7 activations down to 8 bits. As for the weights, while most of them need 12 bits to be well covered, FC8 weights work well with 6 bits. By these means, we can achieve a 55.64\% saving for weights' storage and 69.17\% for activations' memory traffic. The top1 and top5 accuracies only decrease 2.5\% and less than 1\% respectively, see table 4. 
Since VGG16 itself represents a challenge for standard fixed point due to the activations' range, we can focus on reducing strategically the main key components, see Fig. \ref{fig:layer}-a. Conv1.1 and Conv1.2 layers account for over 25\% of the activations while FC6 contains almost 75\% of the weights. We set all layers' activations to 16 bits while reducing Conv1.1, Conv1.2 to 10 and 12 bits and we keep all weights in floating point except FC6, which is set to 8-bit fixed point. By these means, we obtain a reduction of over 53\% in activations' memory traffic and 55\% in weights' storage with less than 1\% drop in top1 and top5 accuracies, see table 5. 
\subsubsection{\textbf{K-means clustering}}
K-means clustering, using a linear quantizer, approximates better to the baseline implementation than dynamic fixed-point quantization as it keeps the entire range of the data distribution. As we have seen for fixed-point quantization, range is the key factor to achieve good accuracy. By contrast, k-means clustering overcomes this factor but deals with the approximation of all variables sharing common value per interval. In Alexnet's case study, we achieve a reduction of 59.59\% in weights' storage and 75.42\% in activations' memory traffic. Although the L2-Norm distance in the last layer is lower than dynamic fixed point, this reduction is not reflected in the Imagenet test with a drop of 9.12\% and 7.24\% in top1 and top5 accuracy respectively, see table 4. 
VGG16 represents a clear example where k-means clustering outperforms dynamic fixed point due to the data-range key factor. Smaller number of bits can be used to represent a few intervals while keeping the whole data range. Thereby, we quantize all layers' activations to 8 bits (except FC8\footnote{FC8 contains only 1000 activations (negligible compared to other layers) and degrades the performance of the network when it is quantized as slight changes in the last layer can produces variations in the top scores of the Imagenet test.}) and FC6's weights to 6 bits achieving 75\% reduction in activations' memory traffic and 60\% in weights' storage. The top1 and top5 accuracies achieved in this case are 2,6\% and 1,6\% respectively as depicted in table 5.
\begin{figure*}[ht]
\centering
    \includegraphics[width=0.5\textwidth, height=0.6in]{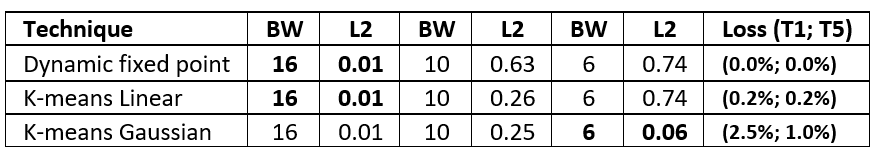}
	\captionof{table}[foo]{Technique comparison when only quantizing FC6 of Alexnet. BW stands for bit width and L2 for the L2-Norm distance between [0, 1] in the last layer. Loss is given for the option in bold.}
    \label{fig:table3}
\end{figure*}
\begin{figure*}[ht]
\centering
    \includegraphics[width=0.7\textwidth, height=1.1in]{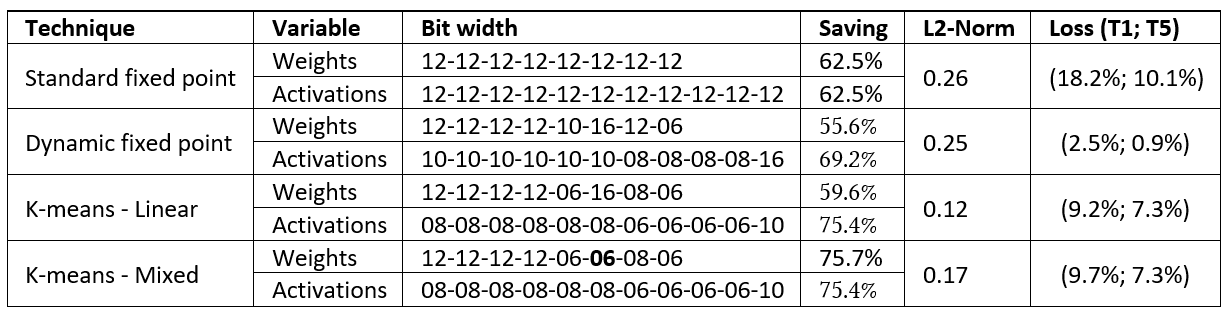}
	\captionof{table}[foo]{Techniques' best results for Alexnet. All percentage results are relative to the baseline floating-point implementation. K-means mixed is equal to k-means linear but only applying a Gaussian quantizer to FC6's weights which are the largest in size. K-means number of clusters is defined as power of two of the layer bit width. L2-Norm is calculated as in Eq. 1.}
    \label{fig:table4}
\end{figure*}
\subsection{Layer distribution}
In this section we argue that a Gaussian quantizer for k-means clustering can obtain better results for certain layers than linear quantizers, if a Gaussian distribution represents better the data distribution, e.g. FC6 in Fig. \ref{fig:distri}-b. On the other hand, other layers e.g. Pool1 are better represented by a linear quantizer as can be seen in Fig. \ref{fig:distri}-c. Gaussian k-means clustering comprises a broad range of the data distribution but not entirely as linear k-means. By contrast, it focuses around the mean and saturates the extremes based on the standard deviation. This configuration improves the precision of the interval and may as well improve the overall accuracy when a single layer is quantized, see table 3. However, when the whole network is quantized using a linear quantizer and we apply a Gaussian quantizer to a specific layer, it does not perform better than applying linear quantizers to all layers. An example is given in table 4 where a Gaussian quantizer is applied to FC6 saving significant memory space and only performing slightly worse than linear quantizer.   
\begin{figure}
\centering
    \includegraphics[width=0.45\textwidth, height=0.9in]{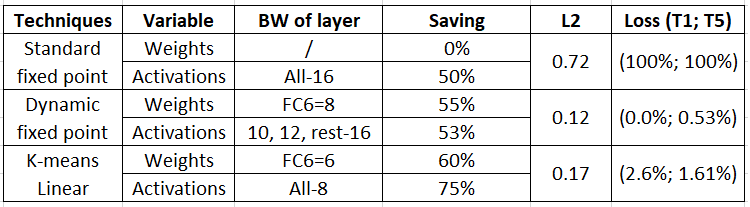}
	\captionof{table}[foo]{Technique comparison when for VGG16. BW stands for bit width and L2 for the L2-Norm distance.}
    \label{fig:table5}
\end{figure}
\subsection{Discussion and comparison}
Dynamic fixed point easily outperforms standard fixed point as it adapts to the data distribution of the network. However, dynamic fixed point may still struggle to bring large memory reductions when wide data ranges appear. In this case, K-means clustering can outperform dynamic fixed point as shown for VGG16. On the other hand, k-means clustering may lack precision since all the variables in an interval share a value which, although being closer to the baseline, produces noise. This noise, in turn, generates a mismatch in the top1 and top5 accuracy test. We believe that the noise could be reduced if fine-tuning is applied as it allows the weights to dynamically adapt to the data path earlier modified by the quantization process.
At last, we show an overview comparison in table 6 between Alexnet's version implemented in this work and Ristretto's (Caffenet) \cite{11}. Dynamic fixed point implemented in this version suffers from having all bit widths equal and it is not shown. By contrast, k-means clustering linear performs well for layer activations quantization and outperforms Ristretto for only FC parameters by using mixed distribution.
\begin{figure}
\centering
    \includegraphics[width=0.4\textwidth, height=0.5in]{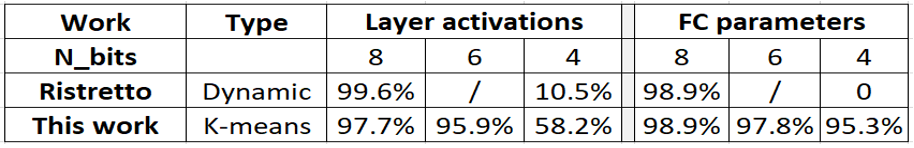}
	\captionof{table}[foo]{Comparison with Ristretto when quantizing only layer activations or FC parameters. Percentage represents top1 relative accuracy respect to each baseline work.}
    \label{fig:table6}
\end{figure}

\section{CONCLUSIONS AND FUTURE WORK}
We have presented an inference engine for quantization and given a comparison for three different techniques when quantization is applied to Alexnet and VGG16. We have shown that dynamic fixed-point greatly improves over standard fixed-point quantization saving over 55\% in memory storage and 69\% in memory traffic, with less than 1\% drop in top5 accuracy for Alexnet. Furthermore, dynamic fix-point quantization can be still improved by k-means clustering since it preserves the range of the data, key factor for quantization, as shown for VGG16 where we obtain 75\% saving in memory traffic and 60\% in weights' storage. We have also demonstrated that Gaussian quantizers for k-means clustering can achieve better performance than linear if the data fits the Gaussian distribution.
Our approach without retraining is complementary to those that perform retraining and avoids the need of massive computational resources and expertise which may be impractical for many users. Furthermore, the engine that we propose brings rich information about the network which can be used to improve retraining. In this work, the authors have focused on the analysis of direct quantization introspectively and retraining is the obvious path to follow next as we expect it to increase the accuracy of the quantized networks, especially for k-means clustering.

\begin{acks}
This project has received funding from the European Union's Horizon 2020 research and innovation programme under grant agreement No 732204 (Bonseyes). This work is supported by the Swiss State Secretariat for Education, Research and Innovation (SERI) under contract number 16.0159. The opinions expressed, and arguments employed herein do not necessarily reflect the official views of these funding bodies.
\pagebreak
\end{acks}